\documentclass[10pt, letter,twocolumn]{IEEEtran}

\usepackage{amsmath}
\usepackage{gensymb}
\usepackage{balance}
\usepackage{float}
\usepackage{romannum}

\usepackage{graphicx}
\usepackage{comment}
\usepackage{subcaption}
\usepackage{algorithm}
\usepackage{algorithmic}
\usepackage{amsmath}
\usepackage{amssymb}
\usepackage{amsxtra}
\usepackage{multirow}
\usepackage{mathtools}
\usepackage{cite}	
\usepackage{color}
\usepackage{bbm}

\title{A Machine Learning Smartphone-based Sensing for Driver Behavior Classification\vspace{-0.2cm}} 
\author{\IEEEauthorblockN{\large Sarra Ben Brahim$^{1}$, Hakim Ghazzai$^{2}$, Hichem Besbes$^{1}$, and Yehia Massoud$^{2}$}\\\IEEEauthorblockA{\small $^{1}$Higher School of Communications of Tunis, University of Carthage, Tunis, Tunisia\\
$^{2}$CEMSE Division, King Abdullah University of Science and Technology (KAUST), Thuwal, Saudi Arabia
}
\vspace{-0.5cm}}

\begin{document}
\maketitle
\thispagestyle{empty}
\begin{abstract}
\boldmath
Driver behavior profiling is one of the main issues in the insurance industries and fleet management, thus being able to classify the driver behavior with low-cost mobile applications remains in the spotlight of autonomous driving.
However, using mobile sensors may face the challenge of security, privacy, and trust issues.
To overcome those challenges, we propose to collect data sensors using Carla Simulator available in smartphones (Accelerometer, Gyroscope, GPS) in order to classify the driver behavior using speed, acceleration, direction, the 3-axis rotation angles (Yaw, Pitch, Roll) taking into account the speed limit of the current road and weather conditions to better identify the risky behavior.
Secondly, after fusing inter-axial data from multiple sensors into a single file, we explore different machine learning algorithms for time series classification to evaluate which algorithm results in the highest performance. 
\end{abstract}
\vspace{-0.2cm}
\begin{IEEEkeywords}
Artificial intelligence, intelligent transportation systems, driver profiling, time series classification. 
\end{IEEEkeywords}

\vspace{-0.3cm}

\section{Introduction}
Over the last two decades, Road Traffic Accidents (RTAs) are increasingly being recognised as a growing public health problem. A statistical projection of traffic fatalities for 2020 from the National Highway Traffic Safety Administration (NHTSA) shows that an estimated 38,680 people died in motor vehicle traffic crashes on U.S. roadways, an increase of 7.2 percent compared to the previous year~\cite{1230}. Aggressive driving is one of the major causes of traffic crashes, according to the NHTSA, which listed impaired driving, over-speeding and sudden lane change as ones of main behaviors causing fatal accidents~\cite{1236}. Therefore, urgent solutions to control road traffic and monitor driving behavior are becoming a necessity for many countries to cope with RTAs~\cite{smartcities3020018}.

Road safety systems and strategies have been developed to safeguard road users and avoid inevitable errors that they might make. They aim to reduce risky behaviors and identify the factors of accidents if they occur. Another approach to promote safe driving is adopted by insurance companies. Premium plans such Usage-Based Insurance (UBI) or Pay-How-You-Drive (PHYD) are proposed to customers to reduce automobile insurance costs by rewarding those who adopt safe driving behavior~\cite{1237}. Nevertheless, tracking the driving behavior in real-time and providing feedback on how a driving is behaving on road is not a straightforward task. Traditional techniques such as the ones using exclusively GPS data to identify the driver behavior are not always effective~\cite{97123}. However, Artificial Intelligence (AI) and the Internet-of-Things have shown promising results in tackling this complex task~\cite{977788,977766,78956,977744,9197640}.

Two categories of sensors are usually involved to monitor the driving behavior: vehicle dynamics and driver dynamics~\cite{977788}. Vehicle dynamics mostly rely on the signals of sensors, such as Global Positioning System (GPS), accelerometers, gyroscopes, and magnetometers. On the other hand, driver dynamics mostly rely on the signals of active sensors, such as video-cameras located inside the vehicle that observe the driver’s behavioral biometrics. Using vehicle dynamics, more precisely smartphone sensors, for driving behavior analysis is the focus of this study due to the fact it has been shown that it is more suitable to use than CAN-bus sensors for driver behavior classification~\cite{977766}. Furthermore, we are looking to design reliable and low-cost solution. 
\begin{figure}[t!]
\centering
 \includegraphics[width=8.5cm]{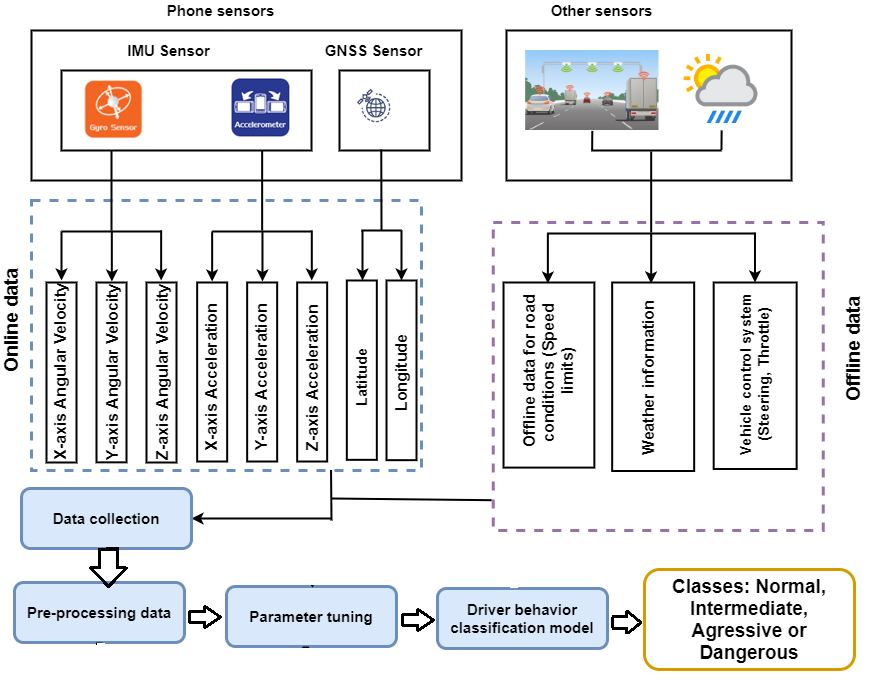}\vspace{-3mm}
\caption{Proposed workflow of the AI-based driver behavior classification.} \vspace{-5mm}
\label{fig1}
\end{figure}

Recently, the authors of~\cite{778899} have conducted a complete literature review on existing studies and techniques about various sensors, technologies, data, and machine learning algorithms used to investigate driver behavior. They also suggested a conceptual framework that uses smartphones to identify the driver behavior. Another study utilized UAH-DriveSet which is a public dataset that provides a wide range of smartphone sensors (e.g., GPS, accelerometers) and even video recordings to distinguish between three driving behaviors; normal, aggressive, and drowzy using Long Term Short Memory (LSTM) and LSTM-Fully Convolutional Network~\cite{977744,977788}. In order to examine the performance of different types of Recurrent Neural Networks (RNNs), the authors of~\cite{232323} used only data that can be obtained from the smartphone's sensors, such as acceleration and GPS. They gathered information from four distinct car routes driven by two different drivers in identical weather conditions. The RNNs were evaluated using seven different driving events, and it was shown that the Gated Recurrent Unit (GRU) outperformed the LSTM and simpleRNN methods for this dataset.





In this paper, we propose to classify the driving behavior into four different classes (normal, intermediate, aggressive, and dangerous) in various external conditions (speed limits, weather conditions, traffic signs) using data collected from smartphones' sensors only such as accelerometer and gyroscope. The data is collected in the form of time series which are analyzed to train an AI-based classifier. Two category of classifiers are investigated including three Gradient Boosting Decision Trees (GBDT) machine learners and three LSTM deep learners. Unlike previous work which used one major dataset, namely the UAH-DriveSet which considers only three behavior classes (normal, drowzy and aggressive) on two different routes, we propose to simulate the driving environment on the Carla simulator and collect the time series data for training and testing. Selected results compare the different implemented methods and show an effective ability to detect the driving behavior using limited sensor sources.


\section{Driver Behavior Classification Method}
As showcased in Fig.~\ref{fig1}, the proposed method encompasses four phases: i) data collection, ii) data pre-processing, iii) parameter tuning, and iv) machine learning model design. 

In this study, we propose to extract information about the driver's risky behavior from the smartphone sensors only as they are equipped with reliable and low-cost sensors that are available with most Android and IOS operating systems such as Global Navigation Satellite System (GNSS), accelerometers, and gyroscopes. 
To achieve a more complete and accurate movement tracking of the vehicle, sensor fusion is therefore required. Collected signals can be detected at unequal intervals so resampling them is an essential step, which will be investigated in the pre-processing phase of the framework. 

Once data is processed, we obtain time series of data points resampled at the same frequency. Before feeding them to the classification model, we perform a parameter tuning to find the best combination of hyperparameters for the machine learner that results in the most accurate classification performance. Finally, we implement a machine learning algorithm for the classification task. In this paper, we propose a comparative study between different machine learners that are the most suitable for the driving behavior classification.

For a better understanding of the driver behavior, we distinguish between four types or classes of the driver behavior:\\
$\bullet$ Normal: the driver is respecting the speed limit given weather conditions and that he/she is safe for other drivers.\\
$\bullet$ Intermediate: the driver is showing few aggressive actions on the road but he/she is not putting other drivers in danger.\\
$\bullet$ Aggressive: the driver is over-speeding, over-steering, weaving in and out of traffic.\\
$\bullet$ Dangerous: the driver is performing many dangerous actions, e.g., dangerous steering, dangerous over-speeding, with a high risk of causing a crash.

This study does not assume that a driver behavior will remain the same over the whole trip. Indeed, we aim to detect different actions performed by the drivers during the same trip and classify them accordingly, e.g., every one minute.

\begin{table}
\parbox{.4\linewidth}{
\caption{Impact of weather on the speed limit}
\centering
\begin{tabular}{ |c|c|  }
 \hline
\textbf{Weather}  & \textbf{Speed reduc-  } \\ [1ex] 
\textbf{type}  & \textbf{tion factor} \\ [1ex] 

 \hline\hline
Sunny & 0\%   \\
 \hline
Soft rain  & 15\%  \\
\hline
Foggy     & 30\% \\
\hline
 Stormy 	 & 40\% \\ 

 \hline
\end{tabular}

\label{tab:weather_impact}
}
\hfill
\parbox{.55\linewidth}{
\caption{Degree of severity of the speeding}
\centering
\begin{tabular}{ |c|c|  }
\hline
 
\textbf{Speed}  & \textbf{Degree of}  \\ [1ex] 
\textbf{Behavior}  & \textbf{aggressiveness }  \\ [1ex] 
 \hline
  \hline
Under speed limit & Normal   \\
 \hline
Overspeed [10,33]\%  &Intermediate  \\
\hline
Overspeed [33,66]\%   & Aggressive\\
\hline
Overspeed 66\%+  & Dangerous  \\ 
 \hline
\end{tabular}\label{tab:speed_distribution}
}\vspace{-3mm}
\end{table}
\begin{table}[t!]
\centering
\caption{Degree of severity of the steering}
\begin{tabular}{ |p{3cm}|p{3cm}|p{1.5cm}|  }
\hline

\textbf{Allowed number of waving in and out
per minute}  & \textbf{Absolute value  of the steering impulse }&\textbf{Steering behavior} \\
\hline
 \hline
Under 5  & Less than 0.15& Intermediate  \\
 \hline
Between 5 and 10   &Between 0.16 and 0.45& Aggressive\\
 \hline
Over 10	 & More than 0.45& Dangerous \\ 
\hline
\end{tabular}

\label{tab:steering_behavior}\vspace{-3mm}
\end{table}
\section{Driver Behavior Classification Phases}
\subsection{Data collection}
In this phase, we define the features considered in the driver behavior classification. Then, we identify the rules of modeling the driver behavior classes and how they will be labeled.

\subsubsection{Feature description}
Data collection in driver behavior analysis can be performed either in real world environment or in a virtual environment. In the first case, although the collected data is more close to the reality, it requires more time and financial effort to gather a dataset. In the second case, driver behavior can be simulated on powerful software that allows a faster collection of data, while providing a full control on the investigated scenarios, e.g., multiple type of roads, traffic situation, and weather conditions. In this study, we perform the data collection using the Carla simulation that we will be discussed further in the next section.

As mentioned earlier, we maintain the gyroscope and the accelerometer to identify the risky behavior~\cite{232323}. Hence, those motion sensors are considered to measure acceleration and rotational forces along three axes. The magnitude of the acceleration is also an important feature that will be exploited with lateral and longitudinal accelerations to detect aggressive driving behavior~\cite{12345678}. The GNSS supplies the longitude and latitude of the vehicle. Steering angle and throttle are also considered in our model since they can reflect a certain degree of aggressiveness~\cite{1234567}. Moreover, every trajectory is enriched with weather information (e.g., rain, fog, humidity) and road data (e.g., traffic light location and speed limits) which are provided by environmental sensors and an offline map available in the smartphone.

All features, described in Fig.~1, will be gathered in one output as time-series data collected from the vehicle trip.

\subsubsection{Selected Rules about Modeling the Driver Behavior}
Over speeding is considered as one of the major factors contributing to severe car crashes on highways and motorways. Thus, speed limits set by local authorities represent the most popular countermeasure. 
We summarize, in Table~\ref{tab:weather_impact}, the impact of the weather on the speed after considering the reduction of visibility and pavement friction due to bad weather conditions. On the other hand, based on California speeding tickets fees that vary from the range of 35\$ to penalize an overspeed between 1.5 km/h and 25 km/h to 200\$ for an overspeed of 160 km/h or more, we have constructed, in Table~\ref{tab:speed_distribution}, the degree of aggressiveness of the driving based on the speed behavior. Due to the fact that cutting off other motorists in traffic is dangerous and reckless behavior, in Table~\ref{tab:steering_behavior}, we present some examples of aggressive steering based on our approach, assuming, for simplification, that the driver sessions are recorded on highways only and that there is no turning, and the steering angle is between 1 and -1 (maximum right and left steering).



\subsection{Data Pre-processing}
We propose to pre-process the collected data before initiating the driver behavior analysis.
\subsubsection{Preparation step}
Data captured from different sensors are measured at irregular intervals due to latency or other external factors. Hence, resampling the time series data is essential as it involves changing the frequency of our time series observations so that it can be at the same frequency.
In our case, resampling will occur every 250 milliseconds so all captured signals from sensors will be available every 250 milliseconds which is sufficient since not major events from driving sessions can happen during that period, e.g., sudden acceleration, sudden deceleration, over-steering can be detected.

\subsubsection{Augmentation phase and normalization}
In this driving behavior study, we resort to the sliding window principle as a feature engineering tool to achieve more accurate results in the classification. Actually, sliding windows based algorithms are widely used for time-series segmentation, in order to reveal the underlying properties of the signals. Previous studies using the sliding window principle in driver behavior classification~\cite{977788,977766,977744} have shown that it provides more accurate results. The sliding window principle means dividing the original time series into multiple equal length subsequences to give more insights about the data and increase the number of samples.


We choose to employ a 1-minute sliding window with 50\% overlap. Each sliding window is then labeled according to the four potential driver behavior classes. Then, all sliding windows are assembled into a single dataset. Since our data contains multiple measurements from different sensors, we proceed with its normalization and shift and rescale the collected features to the same range. 


  
   
          
\vspace{-0.4cm}
\subsection{Parameter Tuning}
Parameter tuning allows to customize the classification models so that they generate the most precise outputs.
For the same machine learning model, different parameters are required to control the convergence speed and avoid overfitting. For instance, in gradient boosting decision trees algorithms, some parameters should be tuned to get the best combinations that give better results in the driver behavior classification such as the number of trees, the learning rate, and the tree depth.
For deep learning algorithms, we choose to optimize the number of epochs and batch size. 

\vspace{-0.4cm}
\subsection{Machine Learning Models}
In this study, we investigate six classification models to identify the driver behavior and distinguish between the four potential driving behavior classes.
The first category is the GBDT, which combines multiple weak learners, typically a decision tree, to form a stronger model that outperforms the base model. When a boosting decision tree is added, it learns from the error of previous individual tree. All the trees are connected in series and each tree tries to minimize the error of the previous tree. Therefore, GBDT uses an incremental learning process as it learns from the mistake of each individual weak learner. Due to this sequential connection, boost algorithms are generally slow to learn, but show great performance on the classification task. Yet, different algorithms for gradient boosting on decision trees, e.g., CatBoost, XGBoost, and LightGBM, have proven their efficiency in the multi-classification task in terms of accuracy and training time~\cite{0123}.

Similar to GBDT, we propose to classify the driver behavior with three different architectures of LSTM. The first model is constructed based on a custom
design of LSTM architecture consisting of one LSTM cell layer called Stacked-LSTM and organized as a many-to-one structure since our dataset contains multiple different trips with fixed duration. The second model is a combination of a Convolutional Neural Network (CNN) and LSTM, defined as CNN-LSTM. 
It can be seen as two sub-models: the CNN model (on the front) is used feature extraction and then followed by the LSTM layers to interpret features across time steps. 
In the case of this study, the input will be subsequences equal in length instead of one sequence. Each will be considered for the CNN layer as a block. CNN will use those blocks to detect common patterns for each feature. This technique enables the identification and analysis of the patterns of each of the subsequence and hence, learn in a more efficient way the temporal features to achieve better classification results using LSTM.


The third model is a combination of a Fully Convolutional Network (FCN) and an LSTM. It is denoted as FCN-LSTM and utilizes temporal convolutions layers as feature extractors. The LSTM-FCN model achieves the state-of-the-art in many sequence classification problems and has received a lot of attention from the time series classification community. The ability of these models to compute features on their own presents a significant advantage, since they eliminate the need for extensive domain expertise and manual feature extraction. Furthermore, both of these models are easily scalable to huge amounts of time series data generated daily by automated procedures.  
The FCN block consists of three convolution layer followed by a batch normalization, and an activation function. Following the final convolution block, a global average pooling is used to reduce the number of parameters in the model before classification. For the LSTM block, it consists of one layer of LSTM which is followed by a dropout to prevent overfitting. The outputs of the global pooling layer and the LSTM
block are concatenated and fed into a softmax classification layer which will predict the corresponding class.


 

Thus, we propose a comparative study between gradient boosting algorithms (Catboost, XGBoost, LightGBM) and different architectures of LSTM (LSTM, CNN-LSTM, FCN-LSTM) with the aim of achieving effective classification performance, specifically the macro F1-score, which considers false alarms and imbalanced class distribution.

\vspace{-0.3cm}
\section{Results and Discussions}

\subsection{CARLA Simulator and Dataset Generation}
\begin{figure}[t]
  \centering
 \includegraphics[width=7cm]{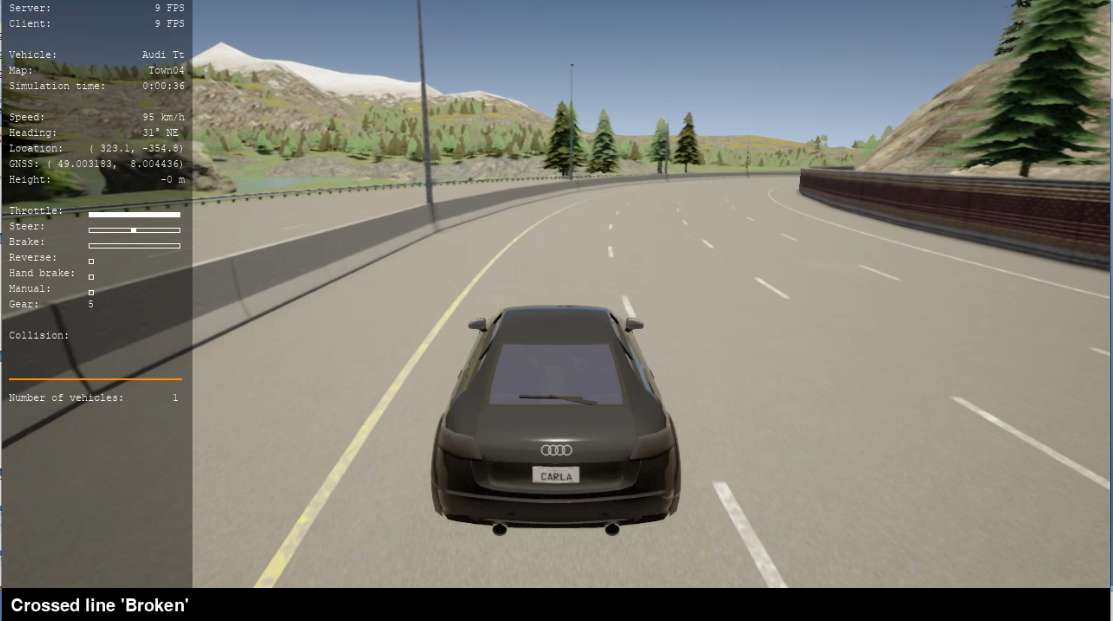}
\caption{Smartphone sensors collected using the Carla simulator.}
\vspace{-4mm}
\label{fig3carla}
\end{figure}

After modeling multiple driving behaviors and generating different scenarios of trips, we collect more than 500 minutes using the Carla simulator as shown in Fig.~\ref{fig3carla}~\cite{Dosovitskiy17,9345899,9439506}. The driving sessions are recorded on four different towns with two types of road (motorway and secondary) with four different weather conditions (sunny, soft rain, cloudy \& foggy, and stormy) and four styles of driving (normal, intermediate, aggressive, and dangerous) that occur along the trips such that a reasonable balanced dataset between the different classes is generated.

 \subsection{Selected Results}
In this section, we compare the performances of the CatBoost, the XGBoost, and the LightGBM to evaluate their efficiency in driver behavior classification. 
In our evaluation, we consider the following metrics:
\begin{equation}
      \text{Average Accuracy}= \frac{1}{N}\sum\limits_{\substack{i=0}}^{N}\frac{{TP_i}+{TN_i}}{{TP_i}+{TN_i}+{FP_i}+{FN_i}},  \hspace{-0.1cm}  
            \label{Accuracy}
          \end{equation}
        \begin{equation}
      \text{Macro  Precision}= \sum\limits_{\substack{i=0}}^{N} \text{Precision}_i= \frac{1}{N}\sum\limits_{\substack{i=0}}^{N} \frac{{TP_i}}{{TP_i}+{FP_i}},  \hspace{-0.1cm}  
            \label{precision}
          \end{equation}
\begin{equation}
        \text{Macro Recall}=\sum\limits_{\substack{i=0}}^{N} \text{Recall}_i= \frac{1}{N}\sum\limits_{\substack{i=0}}^{N}\frac{{TP_i}}{{TP_i}+{FN_i}}, 
            \label{Recall}
          \end{equation}
 \begin{equation}
        \text{Macro F1-score}=\frac{1}{N}\sum\limits_{\substack{i=0}}^{N}\frac{\text{Precision}_i\;\text{Recall}_i}{\text{Precision}_i+\text{Recall}_i},
            \label{F1-score}
          \end{equation}
where $N$ is the number of classes which is equal to four in our case, $TP_i$, $FP_i$, $TN_i$, and $FN_i$ are the true positives, false positives, true negatives, and false negatives per class. We use macro-averaging technique when calculating the accuracy, recall, precision and F1-score to cope with the problem of imbalanced data.

In Table~\ref{tab:parameter_tuning}, we present the best combinations of hyperparameters as a result from the random search technique which tests different parameters including the number of trees, the learning rate, the tree depth, and the L2-regularization in addition to the used loss function.
\setlength{\tabcolsep}{4pt}     

\begin{table}[t!]
\caption{Results of the hyperparameter optimization}
\centering
\begin{tabular}{ |p{1cm}|p{2.25cm}|p{2.25cm}|p{2cm}|  }
\hline

\textbf{Training parameters}  & \textbf{CatBoost }&\textbf{XGBoost } &\textbf{LightGBM}  \\
\hline
 \hline
\textbf{Loss function } & 'MultiClass' & 'mlogloss' &  'multi-logloss'\\
 \hline
\textbf{Best parameters } &  learning rate=0.73,
              \newline
              max depth=8,
              \newline
              n estimators=88,
              \newline
              random state=10
              &  
              learning rate=0.28,
              \newline
              max depth=4,
              \newline
              n estimators=13,
              \newline
              random state=10,
              &  
               learning rate=0.04, 
               \newline
               max depth=8,
                      \newline
               n estimators=53,
                      \newline
                random state=10\\
 \hline

\hline
\end{tabular}

\label{tab:parameter_tuning}\vspace{-3mm}
\end{table}

\begin{table}[t!]
\caption{Comparison between the GBDT algorithms}
\centering
\begin{tabular}{ |p{2cm}|p{1.5cm}|p{1.5cm}|p{1.5cm}|  }
\hline

\textbf{Metrics}  & \textbf{CatBoost }&\textbf{XGBoost } &\textbf{LightGBM}  \\
\hline
 \hline
\textbf{Average Accuracy } & 80\%  & 82\% & 88\% \\
 \hline
\textbf{Macro Precision} & 83\%  &83\%  &88\%  \\
           
 \hline
\textbf{Macro Recall} &81\% &83\%&89\% \\

 \hline
\textbf{Macro F1-score} & 81\% &  83\% &88\% \\
 \hline
\end{tabular}
\label{tab:metrics_classification}\vspace{-3mm}
\end{table}

Based on the comparison between the three different gradient boosting algorithms described in Table~\ref{tab:metrics_classification}, and after splitting our dataset into 70\% of training and 30\% of testing, we find that LightGBM has achieved an F1-score of 88\% with testing sets. Nore that, XGBoost is the fast in training taking only few seconds to choose the best parameters and to classify the behavior of the driver.

In Table~\ref{tab:LSTM metrics}, we test different numbers of epochs and batch sizes before concluding that training on 150 epochs and a batch size of 20 is the best configuration for our three models due to the fact that we have 500 samples.
The table evaluates the performance of the different LSTM architectures on the testing set to identify which architecture achieves the best F1-score and thus the least false positive and false negative rates. We conclude that the best architecture among LSTM is FCN-LSTM which achieves an F1-score of 74\% which is higher than LSTM that achieves 58\% and CNN-LSTM 68\%. Indeed, FCN-LSTM are effective in sequence classification and hence, more suitable for driver behavior classification.
\begin{table}[t!]
\caption{Performance of the different LSTM architectures}
\centering
\begin{tabular}{ |p{2cm}|p{1.75cm}|p{1.75cm}|p{1.75cm}|  }
\hline

\textbf{Training parameters}  & \textbf{LSTM }&\textbf{CNN-LSTM } &\textbf{LSTM-FCN}  \\
\hline
 \hline
\textbf{Loss function } & 'categorical
\newline
crossentropy' & 'categorical
\newline
crossentropy' & 'categorical
\newline
crossentropy'\\
 \hline
\textbf{Optimizer}& Adam & Adam &Adam\\
 \hline

 \hline
 \textbf{Average accuracy}&70\%&76\%& 79\%\\
 \hline
 
 \textbf{Macro Precision}&68\%&79\%& 75\%\\
 \hline
 \textbf{Macro recall}&60\%&69\%& 74\% \\
 \hline
 \textbf{Macro F1-score}&58\%&68\%& 74\%\\

\hline
\end{tabular}
\vspace{-0mm}
\label{tab:LSTM metrics}
\end{table}

\begin{figure}[tb!]
\centering
 \includegraphics[width=8.5cm]{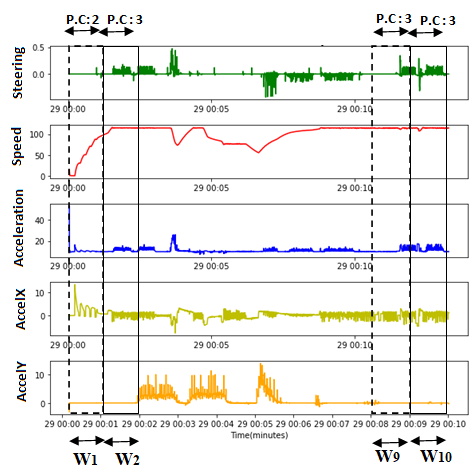}\vspace{-2mm}
\caption{Example of test data collected from the simulator using the LightGBM classifier.}
\vspace{-5mm}
\label{fig5:testing}
\end{figure}

Fig.~\ref{fig5:testing} shows how the proposed solution, in this case the LightGBM, can classify the driver behavior in practice. In this example, we record a ten-minute trip. The classifier observes this trip as ten sub-trips of one-minute duration using the sliding window principle and aims to predict the corresponding class. Fig.~\ref{fig5:testing} illustrates one test example of the simulated data with $W_j$ denotes the $j^\text{th}$ sliding window and "P.C" refers to the predicted class with 0 for normal behavior, 1 for intermediate, 2 for aggressive and 3 for dangerous. The figure illustrates the most influential features: steering, speed (km/h), and the three acceleration components. In this testing phase, the predicted class of $W_2$ is 3, which can be justified by the detected over speeding noticeable specifically in the AccelX. This means if one of the features is detected abnormal given the weather condition and the road status, the whole sub-trip can be classified as dangerous. The predicted classes of those sub-trips can be further analyzed to get a complete overview of the driver behavior along its whole journey.

\section{Conclusion}
In this paper, we have investigated the driving behavior classification problem using smartphone-based sensing only.
After collection of our time series data under different road and weather conditions, we fed our simulated data into our AI-agent to detect different driving behavior classes and to identify every driver profile. Results have shown the ability of both machine and deep learning models to achieve an accuracy greater than 88\% in detecting driving profile for a one-minute duration trip. 


This framework is a low-cost solution for many fleet management and monitoring problems and can further be used to prevent cars from accidents. 
Although the collected data from simulations mimics real-life situations in terms of weather conditions, types of aggressive behaviors, in the future extension of this work, we aim to focus on collecting realistic data and provide more generalized driver behavior classification framework that considers other road types other than highways.


\bibliographystyle{IEEEtran}\balance

\bibliography{References}

\begin{thebibliography}{10}
\providecommand{\url}[1]{#1}
\csname url@samestyle\endcsname
\providecommand{\newblock}{\relax}
\providecommand{\bibinfo}[2]{#2}
\providecommand{\BIBentrySTDinterwordspacing}{\spaceskip=0pt\relax}
\providecommand{\BIBentryALTinterwordstretchfactor}{4}
\providecommand{\BIBentryALTinterwordspacing}{\spaceskip=\fontdimen2\font plus
\BIBentryALTinterwordstretchfactor\fontdimen3\font minus
  \fontdimen4\font\relax}
\providecommand{\BIBforeignlanguage}[2]{{%
\expandafter\ifx\csname l@#1\endcsname\relax
\typeout{** WARNING: IEEEtran.bst: No hyphenation pattern has been}%
\typeout{** loaded for the language `#1'. Using the pattern for}%
\typeout{** the default language instead.}%
\else
\language=\csname l@#1\endcsname
\fi
#2}}
\providecommand{\BIBdecl}{\relax}
\BIBdecl

\bibitem{1230}
{NHTSA Media}, in \emph{2020 Fatality Data Show Increased Traffic Fatalities
  During Pandemic}, June 3, 2021 | Washington, DC.

\bibitem{1236}
{U.S. Department of Transportation, National Highway Traffic Safety
  Administration.}, in \emph{Facts + Statistics: Aggressive driving}, 2019.

\bibitem{smartcities3020018}
M.~C. Lucic, X.~Wan, H.~Ghazzai, and Y.~Massoud, ``Leveraging intelligent
  transportation systems and smart vehicles using crowdsourcing: An overview,''
  \emph{Smart Cities}, vol.~3, no.~2, pp. 341--361, May 2020.

\bibitem{1237}
{M. Carfora}, { F. Martinelli}, { F. Mercaldo}, {F. Nardone}, {V. Orlando}, {A.
  Santone}, and {G. Vaglini}, ``A “pay-how-you-drive” car insurance
  approach through cluster analysis,'' in \emph{Soft Computing.}, 2019.

\bibitem{97123}
{Grengs, J.}, {Wang, X.}, and {Kostyniuk, L.}, ``Using gps data to understand
  driving behavio,'' in \emph{Journal of Urban Technology}, 19 Jan 2009, pp.
  33--53.

\bibitem{977788}
{K. Saleh}, { M. Hossny}, and {S. Nahavandi}, ``Driving behavior classification
  based on sensor data fusion using $lstm$ recurrent neural networks,'' in
  \emph{2017 IEEE 20th International Conference on Intelligent Transportation
  Systems (ITSC)}, October 2017.

\bibitem{977766}
{V. Vaitkus}, { P. Lengvenis}, and {G. Žylius}, ``Driving style classification
  using long-term accelerometer information,'' in \emph{2014 19th International
  Conference on Methods and Models in Automation and Robotics (MMAR)},
  September 2014.

\bibitem{78956}
{A. Abdelrahman}, , {H, S. Hassanein}, and { N. Abu Ali}, ``Robust data-driven
  framework for driver behavior profiling using supervised machine learning,''
  in \emph{IEEE Transactions on Intelligent Transportation Systems}, November
  2020.

\bibitem{977744}
{Y. Moukafih}, { H. Hafidi}, and {M. Ghogho}, ``Aggressive driving detection
  using deep learning-based time series classification,'' in \emph{2019 IEEE
  International Symposium on INnovations in Intelligent SysTems and
  Applications (INISTA)}, July 2019.

\bibitem{9197640}
X.~Wan, M.~C. Lucic, H.~Ghazzai, and Y.~Massoud, ``Empowering real-time traffic
  reporting systems with nlp-processed social media data,'' \emph{IEEE Open
  Journal of Intelligent Transportation Systems}, vol.~1, pp. 159--175, 2020.

\bibitem{778899}
{F. Lindow} and {A. Kashevnik}, ``Driver behavior monitoring based on
  smartphone sensor data and machine learning methods,'' in \emph{Conference of
  Open Innovations Association (FRUCT'19)}, 2019.

\bibitem{232323}
{J. Ferreira Ju´nior}, {E. Carvalho}, {B, V. Ferreira}, {C. de Souza}and {Y.
  Suhara}, {A. Pentland}, and {G. Pessin}, ``Driver behavior profiling: An
  investigation with different smartphone sensors and machine learning,''
  April,2017.

\bibitem{12345678}
{ S. Chigurupati}, { S. Polavarapu}, {Y. Kancherla}, and { A. Kousar Nikhath4},
  ``Integrated computing system for measuring driver safety index,'' in
  \emph{International Journal of Emerging Technology and Advanced Engineering},
  Juin 2012.

\bibitem{1234567}
{ C. G. Quintero M.}, , {J. Oñate López}, and { A, C. Cuervo Pinilla},
  ``Driver behavior classification model based on an intelligent driving
  diagnosis system,'' in \emph{International IEEE Conference on Intelligent
  Transportation Systems}, 25 October 2012.

\bibitem{0123}
{J. Tanha}, { Y. Abdi }, {N. Samadi}, { N. Razzaghi }, and {M. Asadpour },
  ``Boosting methods for multi-class imbalanced data classification: an
  experimental review,'' in \emph{Journal of Big Data}, November 2020.

\bibitem{Dosovitskiy17}
A.~Dosovitskiy, G.~Ros, F.~Codevilla, A.~Lopez, and V.~Koltun, ``{CARLA}: {An}
  open urban driving simulator,'' in \emph{Proceedings of the 1st Annual
  Conference on Robot Learning}, 2017.

\bibitem{9345899}
H.~Friji, H.~Ghazzai, H.~Besbes, and Y.~Massoud, ``A dqn-based autonomous
  car-following framework using rgb-d frames,'' in \emph{2020 IEEE Global
  Conference on Artificial Intelligence and Internet of Things (GCAIoT)}, 2020,
  pp. 1--6.

\bibitem{9439506}
M.~Masmoudi, H.~Friji, H.~Ghazzai, and Y.~Massoud, ``A reinforcement learning
  framework for video frame-based autonomous car-following,'' \emph{IEEE Open
  Journal of Intelligent Transportation Systems}, vol.~2, pp. 111--127, 2021.

\end{thebibliography}
\end{document}